\pdfoutput=1

\documentclass[11pt]{article}

\usepackage[]{acl}

\usepackage{times}
\usepackage{latexsym}

\usepackage[T1]{fontenc}

\usepackage[utf8]{inputenc}

\usepackage{microtype}

\usepackage{algorithm}
\usepackage{algorithmic}
\usepackage{amsmath}
\usepackage{mathrsfs} 
\usepackage{graphicx}
\usepackage{caption}
\usepackage{subfigure}
\usepackage{epstopdf}
\usepackage{amsfonts}
\usepackage{color}
\usepackage[algo2e,ruled,vlined]{algorithm2e}
\usepackage{setspace}
\usepackage{booktabs}

\usepackage{amsmath}
\usepackage{makecell}

\usepackage{amsfonts}
\usepackage{booktabs}
\usepackage{multirow}
\usepackage{stfloats}

\newcommand{\eat}[1]{}
\newcommand{\eg}{{\em e.g.,~}}     
\newcommand{\ie}{{\em i.e.,~}}      

\usepackage{enumitem}

\usepackage{newfloat}
\usepackage{listings}

\floatstyle{ruled}
\newfloat{listing}{tb}{lst}{}
\floatname{listing}{Listing}

\title{Prompt-based Conservation Learning for Multi-hop Question Answering}

\author{
Zhenyun Deng, Yonghua Zhu, Yang Chen, Qianqian Qi \\ 
{\bf Michael Witbrock, Patricia Riddle} \\
School of Computer Science, University of Auckland, New Zealand\\
\texttt{\{zden658, yzhu970, beryl413\}@aucklanduni.ac.nz} \\
\texttt{\{yang.chen, m.witbrock, p.riddle\}@auckland.ac.nz} \\
}
        
\begin{document}
\maketitle
\begin{abstract}
Multi-hop question answering (QA) requires reasoning over multiple documents to answer a complex question and provide interpretable supporting evidence. However, providing supporting evidence is not enough to demonstrate that a model has performed the desired reasoning to reach the correct answer. Most existing multi-hop QA methods fail to answer a large fraction of sub-questions, even if their parent questions are answered correctly. In this paper, we propose the Prompt-based Conservation Learning (PCL) framework for multi-hop QA, which acquires new knowledge from multi-hop QA tasks while conserving old knowledge learned on single-hop QA tasks, mitigating forgetting. Specifically, we first train a model on existing single-hop QA tasks, and then freeze this model and expand it by allocating additional sub-networks for the multi-hop QA task. Moreover, to condition pre-trained language models to stimulate the kind of reasoning required for specific multi-hop questions, we learn soft prompts for the novel sub-networks to perform type-specific reasoning. Experimental results on the HotpotQA benchmark show that PCL is competitive for multi-hop QA and retains good performance on the corresponding single-hop sub-questions, demonstrating the efficacy of PCL in mitigating knowledge loss by forgetting.
\end{abstract}

\eat{
Specifically, we first train a model on single-hop QA tasks, then retrain the model to acquire new knowledge from multi-hop QA tasks by freezing the network trained on previous QA tasks and allocating novel sub-networks. Moreover, to condition pre-trained language models (PLMs) to properly stimulate reasoning types required for multi-hop questions, we learn soft prompts in sub-networks to condition frozen trained network to perform question-type-specific reasoning. Experiments on the HotpotQA show that our model achieves a new SOTA, and also has good performance on the corresponding single-hop sub-questions and adversarial HotpotQA.
} 

\section{Introduction}

Multi-hop QA is a challenging task with the goals of reasoning over multiple scattered documents to predict an answer, and providing explanatory supporting evidence \cite{yang-etal-2018-hotpotqa}. By fine-tuning pre-trained language models (PLMs) with task-specific data, most existing multi-hop QA models have achieved good performance in both goals 
\cite{tu2020select,fang2019hierarchical}. 

Despite the success of fine-tuned PLMs on the multi-hop QA task, providing supporting evidence is not enough to demonstrate that a multi-hop QA model has performed the desired multi-hop reasoning to reach the correct answer; it may instead have utilized reasoning shortcuts, having neglected to acquire and retain the single-hop reasoning knowledge essential to reliable interpretability \cite{jiang2019avoiding}. Previous work \cite{tang-etal-2021-multi} has demonstrated that most existing multi-hop QA models with good performance fail to answer a large fraction of the sub-questions whose parent multi-hop questions can be answered correctly. Thus, it is necessary to understand the behaviour on each hop of the reasoning process and mitigate forgetting of the knowledge required for each hop in interpretable multi-hop QA. Doing so should enable humans to better trust the QA mechanism.

\begin{figure}[!t]
	\begin{center}
		{\scalebox{0.65} {\includegraphics{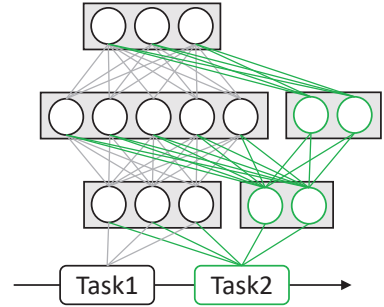}}}
		\vspace{-3mm}
		\caption{\footnotesize An example of conservation learning based on a continual learning mechanism. The neurons on the left are devoted to  Task1 (single-hop QA), and on the right (green) are a novel sub-network created for Task2 (multi-hop QA) that laterally connects to the trained Task1. By adding the sub-network, the model acquires new knowledge of Task2 while retaining knowledge learned in Task1, mitigating forgetting.}
		\label{fig1}
	\end{center}
	\vspace{-5mm}
\end{figure}

In addition, existing QA models integrate all the knowledge by thoroughly pre-training the PLMs on all available data \cite{schwartz2020green}, which integrates the various forms of knowledge from multiple types of questions. However, a downstream QA task may only require knowledge of a specific type. For example, in the multi-hop QA task \cite{yang-etal-2018-hotpotqa}, questions can be roughly divided into two different types: bridging and comparison, each of which requires a specific reasoning strategy to answer. To achieve multi-hop reasoning efficiently, it may be useful for PLMs to disentangle knowledge from other question types and stimulate the appropriate reasoning types required for particular multi-hop questions. 

To address these issues, we propose Prompt-based Conservation Learning (PCL) for multi-hop QA. Specifically: $i$) to train a multi-hop QA model without forgetting, we apply conservation learning based on a continual learning mechanism to acquire new knowledge from multi-hop QA tasks while retaining that previously learned on single-hop QA tasks. As shown in Figure \ref{fig1}, we first train a model on the single-hop QA task; when incorporating the new multi-hop QA task, we freeze the model trained on the single-hop task and expand it by allocating novel sub-networks for new multi-hop knowledge; $ii$) to take full advantage of diverse knowledge in the PLM, we first identify the reasoning type of the multi-hop question as a soft prompt via a transformer-based question classifier, and then transform it into a sub-network that connects laterally with the previously trained QA model, to condition the PLM to perform type-specific reasoning. Since PCL trains the QA model incrementally based on the conserved previously learned parameters, it should be able to perform well on multi-hop QA because it thus retains the previously learned knowledge \cite{parisi2019continual,sun2020ernie}.

Our contributions are summarized as follows:
\begin{itemize} \setlength\itemsep{1pt}
    \item We propose conservation learning for multi-hop QA, which acquires knowledge from the multi-hop QA task while retaining knowledge learned on single-hop QA tasks, which may enable humans to understand the behaviour of each hop in the  reasoning process better.
    \item We propose using a soft prompt based on the reasoning type to condition the PLM, stimulating use of the required knowledge for particular types of multi-hop reasoning.
    \item Our proposed PCL achieves better performance on the HotpotQA leaderboard, while also retaining good performance on the corresponding single-hop sub-questions.
\end{itemize}

\section{Related Work}
\paragraph{Prompt Tuning for PLMs.}
Prompt tuning is an effective mechanism for learning prompts to condition PLMs to stimulate and apply the appropriate knowledge for a specific downstream task \cite{liu2021pre}. \citet{gu2021ppt} propose to initialize soft prompts by adding them into the pre-training stage of few-shot learning. \citet{li2021prefix} prepend a series of learnable continuous embeddings as soft prompts into the input, achieving better performance in text generation tasks. Motivated by these methods, we use the  reasoning types of multi-hop questions as soft prompts to condition PLMs to stimulate the knowledge required to answer multi-hop questions.

\paragraph{Continual Learning for PLMs.}
Continual learning aims to allow systems to repeatedly acquire new knowledge while retaining previously learned experience, mitigating catastrophic forgetting \cite{parisi2019continual}. Conceptually, continual learning can be divided into three categories of technique: $i$) retrain the whole model while imposing additional constraints to retain the important learned model parameters from previous tasks \cite{li2021lifelong}; $ii$) perform memory replay to distill the knowledge from previous model backups \cite{sun2019lamol,rolnick2019experience}; $iii$) freeze the model trained on previous tasks and retrain the model by allocating new neurons or network layers for new tasks \cite{qin2022elle}. In this paper, we propose a learning mechanism based on continual learning, by freezing the model trained on the single-hop QA, and retraining the model for the multi-hop QA using our soft-prompt technique, enables the QA model to achieve single-hop reasoning and multi-hop reasoning simultaneously. Since we only have two tasks, we call this conservation learning. It aims to conserve previously learned knowledge while performing well on a second task; it does not continue for a large number of tasks as in continual learning.

\eat{ In this paper, we apply a single "continual learning" step, by freezing the model trained on the single-hop QA, and retraining the model for our soft-prompt approach to multi-hop QA, enabling the QA model to achieve single-hop reasoning and multi-hop reasoning simultaneously."}

\paragraph{End-to-end Multi-hop QA.}
Existing end-to-end multi-hop QA systems predict the answer and corresponding supporting facts based on the given question and retrieved relevant paragraphs. \citet{qiu2019dynamically}, \citet{fang2019hierarchical}, and \citet{tu2020select} extract information at different levels of granularity as nodes in a graph, and then apply GNN-based methods to answer the question and provide supporting sentences. \citet{shao-etal-2020-graph}, \citet{beltagy2020longformer} and \citet{wu2021graph} argue that graph structures may not be necessary for multi-hop QA, and propose graph-free reasoning models. Unlike these methods, where there is no training requirement for the models to follow the desired reasoning steps to predict the answer, we propose a multi-hop QA framework with separated learning of the intended behaviour of QA models on each hop of the reasoning process and in the final answer.

\begin{figure*}
	\begin{center}
		{\scalebox{0.5} {\includegraphics{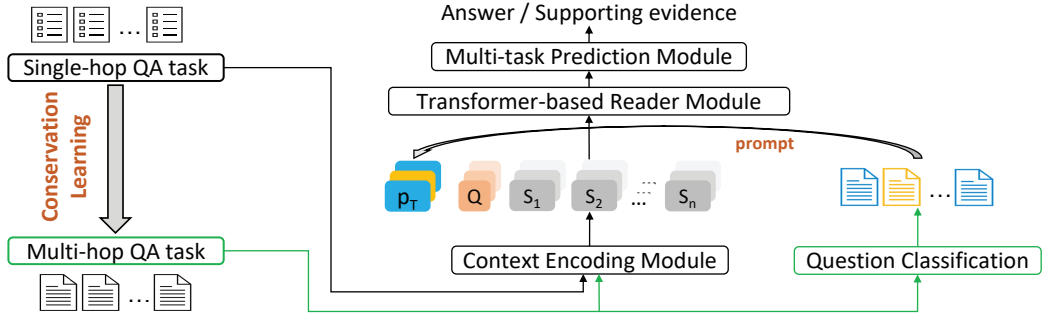}}}
		\caption{\footnotesize
		An overview of our proposed PCL framework for multi-hop QA. Specifically, it involves three key steps, (a) train a QA model to acquire knowledge from single-hop QA tasks; (b) identify reasoning types of multi-hop questions as soft prompts, and transform soft prompts into a sequence of continuous type-specific vectors; (c) retrain the QA model to acquire new knowledge from multi-hop QA tasks by freezing the trained network in single-hop QA task and prepending soft prompt vectors to the input.}
		\vspace{-3mm}
		\label{framework}
	\end{center}
	\vspace{-2mm}
\end{figure*}

\section{Methodology}

\subsection{Overview}
This section, we describe prompt-based conservation learning for multi-hop QA. As illustrated in Figure \ref{framework}, our PCL consists of three components: $i$) we first acquire single-hop QA knowledge by explicitly training on these tasks; $ii$) we then acquire knowledge for the new multi-hop QA task while retaining the learned knowledge using conservation learning; $iii$) we perform type-specific reasoning, identifying the reasoning type of the question via the soft prompt to stimulate application of the appropriate knowledge. 


\subsection{Single-hop QA}
To understand the behaviour of existing QA models on each hop of the reasoning process, we train a QA model based on the PLM, ELECTRA \cite{clark2020electra}  on a single-hop QA task, SQuAD \cite{rajpurkar2016squad}. This QA model contains two modules: context encoding and a transformer-based reader.

\paragraph{Context Encoding.} Given a question $\mathrm{Q}$ and $n$ relevant sentences, we concatenate the question and sentences into an input sequence for the pre-trained ELECTRA encoder to obtain a context representation. Specifically, we formulate the input sequence as ``$\rm [CLS]\ Q\ [SEP]\ yes\ no\ [SEP]\ [SE]\ {s}_1\ [SEP]$ $\rm [SE]\ {s}_2\ [SEP]\ ...\ [SE]\ {s}_i\ [SEP] ...\ [SE]\ {s}_n\ [SEP]$'', where $\rm [SE]$ is a special token delineating supporting evidence, and $\rm yes\ no$  indicates a yes/no answer, which are prepended to the context, subsequently encoded by ELECTRA into the context representation. Consequently, each context sentence $\rm {s}_i$ in the input sequence can interact with other sentences across the concatenated sequence by using a self-attention mechanism; such interactions are crucial for multi-hop QA \cite{zhu2021adaptive}.

\paragraph{Transformer-based Reader.} After context encoding, the context representations are passed through a bi-attention layer to enhance interactions between the question and the context \cite{qiu2019dynamically}. On top of the updated context representation, we have followed \cite{fang2019hierarchical} to design a multi-task prediction module to jointly perform answer and supporting evidence prediction. For answer span prediction, we use two linear layers applied to the context representation to predict the start and end position of the answer. For supporting evidence prediction, we use a binary linear layer to predict a binary relevance label at each sentence start $\rm [SE]$. The final objective is defined as: 
\begin{center}
$\mathcal{L}_{Joint} = \mathcal{L}_{start} + \mathcal{L}_{end} + \lambda_1 \mathcal{L}_{SE}$
\end{center}
where $\lambda_1$ is a hyper-parameter and each loss function $\mathcal{L}$ is the cross-entropy loss between the prediction and ground truth.


\subsection{Multi-hop QA with Conservation Learning}

\paragraph{Origins in Continual Learning.} In one form of Continual Learning, given $\rm N$ existing tasks  $\mathcal{T}_{seq}=\{\mathcal{T}_{0}, \mathcal{T}_{1}, ..., \mathcal{T}_{N}\}$, when a new task $\mathcal{T}_{N+1}$ comes, an additional network is created and the lateral connections with the trained model are learned. To avoid knowledge forgetting, the parameters $\theta^{N}$ learned by the existing tasks $\mathcal{T}_{seq}$ remain unchanged while the new parameter set $\theta^{N+1}$ is learned for the additional network in Task $\mathcal{T}_{N+1}$ \cite{parisi2019continual}. 

\paragraph{Conservation Learning for Multi-hop QA.} To enable a trained single-hop QA model to learn the new knowledge required for a subsequent multi-hop QA task without forgetting previously learned knowledge, we propose a truncated-continual-learning-like method that freezes the learned model and allocates additional sub-networks for the new multi-hop QA tasks. In principle this process could be iterated in continual learning, but here we apply one such step, and coin the term ``conservation learning'' to describe it. PCL's multi-hop QA after conservation learning consists of three components: $i$) question classification: identifying the reasoning type of the multi-hop question; $ii$) paragraph selection: retrieving paragraphs related to the multi-hop question; $iii$) pre-trained soft prompt: conditioning a PLM to perform the type-specific reasoning required for a multi-hop question.

\paragraph{Question Classification.}
Instead of training a separate QA model for each reasoning type, our design uses a single PLM to integrate the knowledge from all reasoning types. To inform this use, we first need to identify the reasoning type of the multi-hop question. Thus, we train a question classifier, also based on ELECTRA, followed by a binary classification layer, to predict the reasoning type for each multi-hop question. The question classifier only takes the question as its input and outputs a relevance score for different reasoning types. The reasoning type with the highest score is selected as the type of multi-hop question.

\paragraph{Iterative Paragraph Selection.} Since not every given paragraph contains relevant information, multi-hop QA models must filter out irrelevant paragraphs. In addition, multi-hop questions also often permit reasoning shortcuts through which QA models can directly locate the final answer by word-matching the question to a single sentence in the paragraph \cite{qi2019answering,qi2020answering}. To discourage this kind of direct but unjustified leap to the answer, we propose to retrieve paragraphs related to the question in an iterative fashion, which encodes the question and previously retrieved paragraphs as a new question vector to retrieve the next relevant paragraph. For simplicity, we use the same model encoder as the question classifier to select relevant paragraphs, except that we take the question $q$ and the paragraph $p$ as the input and output a relevance score for each paragraph. We calculate the score for each paragraph at each retrieval step as follows: 
\begin{equation*}
\small
    \mathcal{P} (P_{seq}|q) = \prod \limits_{t=1}^n \mathcal{P} (p_t | q, p^1, p^2, ..., p^{t-1})
\end{equation*}
where for $t=1$ (\ie the first hop), we only use the original question $q$ for paragraph retrieval. At each subsequent retrieval step, we encode the question $q$ and the most relevant paragraph $p^{t-1}$ in the previous step $t$ as a new question vector to predict the next relevant paragraph. In this way, each subsequent retrieved paragraph is not only related to the question, but also related to the previous retrieved paragraphs, which discourages producing an answer using ``reasoning'' shortcuts and provides a solid basis for multi-hop reasoning in the next step.



\paragraph{Pre-training Soft Prompt.}
To enable the PLM to integrate knowledge from multiple reasoning types, we introduce a soft prompt based on the reasoning type to condition the PLM to perform type-specific reasoning, which is connected laterally to the trained QA model during training. Specifically, we first formulate the input sequence as ``$\rm [CLS]\ Q\ [SEP]\  yes\ no\ [SEP]\ [SE]\ {s}^1_1\ [SEP]\ $ $\rm [SE]\ {s}^2_1\ [SEP]\ ...\ [SE]\ {s}^ {j}_{i}\ [SEP] ...\ [SE]\ {s}^{m}_{n}\ [SEP]$'', where $\rm {s}^{j}_{i}$ indicates the $j$-th sentence in the relevant paragraph $i$; we then utilize the previously trained model to initialize the input sequence to obtain the context representation $\mathbf C = \{\mathbf{c}_0, \mathbf{c}_1, ..., \mathbf{c}_{n-1}\} \in \mathbb{R}^{n \times d}$, where $ n, d$ are the length and the dimension of the context, respectively; we finally transform the reasoning type obtained in the question classification into a continuous trainable vector $\mathbf{p}\in \mathbb{R}^{m \times d}$ and prepend it onto $\mathbf C$, resulting in the new input $\mathbf C' = \{\mathbf{p}_i; \mathbf{c}_0, \mathbf{c}_1, ..., \mathbf{c}_{n-1}\}$, where $m$ is the length of the soft prompt and $\mathbf{p}_i$ is the soft prompt vector of reasoning type $i$. 

Once the new context representation is obtained, it is then processed by the transformer-based reader module. Notably, we optimize $\mathbf{p}_i$ along with other parameters of the PLM during pre-training. During fine-tuning, we prepend the trained soft prompt vector into the input sequence, guiding the model to perform type-specific reasoning. In this way, we condition the PLM to stimulate the proper knowledge required for multi-hop reasoning.

\begin{table*}[!tb]
	\small
	\centering
	\begin{tabular} {l c c c c c c} \hline
		\multirow{2}{*}{Model}  &\multicolumn{2}{c}{Ans} &\multicolumn{2}{c}{Sup} &\multicolumn{2}{c}{Joint} \\ \cmidrule(lr){2-3} \cmidrule(lr){4-5} \cmidrule(lr){6-7} 
		&EM &F1 &EM &F1 &EM &F1  \\\hline
		Baseline Model \cite{yang-etal-2018-hotpotqa} &45.60 &59.02 &20.32 &64.49 &10.83 &40.16  \\
		DecompRC \cite{min2019multi}       &55.20 &69.63 &-     &-     &-     &- \\
		OUNS \cite{perez2020unsupervised}  &66.33 &79.34 &-     &-     &-     &- \\
		QFE   \cite{nishida2019answering}        &53.86 &68.06 &57.75 &84.49 &34.63 &59.61 \\
		DFGN   \cite{qiu2019dynamically}        &56.31 &69.69 &51.50 &81.62 &33.62 &59.82 \\
		SAE-large \cite{tu2020select}     &66.92 &79.62 &61.53 &86.86 &45.36 &71.45 \\
		C2F Reader  \cite{shao2020graph}   &67.98 &81.24 &60.81 &87.63 &44.67 &72.73 \\
		Longformer \cite{beltagy2020longformer}     &68.00 &81.25 &63.09 &88.34 &45.91 &73.16 \\
		HGN-large \cite{fang2019hierarchical}  &69.22 &82.19 &62.76 &88.47 &47.11
		&74.21 \\
		AMGN \cite{liasynchronous}  &70.53 &83.37 &63.57 &88.83 &47.77
		&75.24 \\
		S2G \cite{wu2021graph}  &70.72 &83.53 &64.30 &88.72 &48.60
		&75.45 \\ \hline
		PCL (Ours) &\textbf{71.76} &\textbf{84.39} &\textbf{64.61} &\textbf{89.20} &\textbf{49.27} &\textbf{76.56} \\ \hline
	\end{tabular}
	\caption{\footnotesize Results on the blind test set of HotpotQA in the distractor setting. Our PCL achieves the best performance on the HotpotQA leaderboard. ``-'' denotes the case where no results are available. Leaderboard: https://hotpotqa.github.io/.}
	\vspace{-2mm}
	\label{tab1}
\end{table*}

\section{Experiments}

\subsection{Dataset and Metrics}
We evaluate our model primarily on three datasets: HotpotQA \cite{yang-etal-2018-hotpotqa}, adversarial HotpotQA \cite{jiang2019avoiding} and a manually verified sub-question QA dataset generated from HotpotQA \cite{tang-etal-2021-multi}. To verify whether our PCL can be generalized to other multi-hop QA datasets, we also conduct experiments on two similar datasets: 2WikiMultihopQA \cite{ho2020constructing} and MuSiQue \cite{trivedi2021musique}. Unlike other knowledge-based multi-hop QA datasets \cite{welbl2018constructing,talmor2018web,saxena2020improving} that restrict the final answer to the content of explicit knowledge bases, all QA pairs in the HotpotQA are collected from Wikipedia.

\paragraph{HotpotQA.} Each multi-hop question is provided with ground truth answers and supporting sentences, which enables us to evaluate the performance and interpretability of multi-hop reasoning. There are two reasoning types of questions: bridging and comparison, each of which requires a specific reasoning strategy to answer. 

\paragraph{Sub-question QA dataset.} To analyze whether the multi-hop QA models really perform each hop of the reasoning process, \citet{tang-etal-2021-multi} generate a single-hop sub-question dataset with 1000 manually verified samples for the dev set of HotpotQA for evaluation.

\paragraph{Adversarial HotpotQA.} Multi-hop questions in the HotpotQA often contain reasoning shortcuts through which models can directly find the answer by word-matching the question to a sentence. To avoid this, \citet{jiang2019avoiding} construct adversarial samples by creating contradicting answers to reasoning shortcuts without affecting the validity of the original answers.

\paragraph{Multi-hop QA Dataset.} Unlike HotpotQA, 2WikiMultihopQA evaluates the interpretability of the multi-hop QA model not only with supporting evidence, but also with entity-relation tuples. However, for a fair comparison, we do not use the entity-relation tuples in our training. MuSiQue has richer multi-hop questions with 2-4 hops.

\paragraph{Metrics.}We use Exact Match (EM) and Partial Match (F1) to evaluate the model performance on answer and supporting facts prediction, and a joint EM and F1 score to evaluate the final performance.

\subsection{Implementation Details}
We adopt ELECTRA-large \cite{clark2020electra} as the skeleton for each module. Our released implementation is based on Huggingface \cite{wolf2020transformers}. For question classification and paragraph selection, we train the models for 5 epochs using Adam optimizer, with a batch size of 12, a learning rate of $2\times 10^{-5}$, a warm-up rate of 0.1 and $\ell_2$ weight decay of 0.01. For question answering, we use the same setting as stated above, except for a learning rate of $3\times 10^{-5}$ and an additional prompt length of 2 tokens. The hyper-parameter of $\lambda_1$ is set to 2.
Only the context encoding module is frozen during Conservation Learning and additional weights are added to connect the soft prompts. 


\subsection{Main Results}
We compare our PCL model with other published baselines on the test set of HotpotQA in the distractor setting. As shown in Table \ref{tab1}, we observe that our PCL QA-system outperforms all comparison baselines on every metric and achieves the best performance on the HotpoQA dataset, demonstrating the progress made by PCL in multi-hop QA. Specifically, under the same setting, using a transformer-based ELECTRA model, PCL achieves a 1.12/0.91 improvement on the Joint EM/F1 score, compared with the best graph-free model S2G. This indicates that the effectiveness of the proposed conservation learning and soft prompts. For the best graph-based model AMGN, PCL improves the Joint EM/F1 score by 1.5/1.32, which shows that good performance can be achieved without constructing a graph. In the next section, we provide a detailed analysis to evaluate the performance of conservation learning and soft prompts in our PCL model.

\begin{table*}[!ht]
	\begin{tabular} {m{0.1cm}<{\centering} m{0.4cm}<{\centering} m{0.6cm}<{\centering}| m{0.7cm}<{\centering} m{1.5cm}<{\centering} m{0.6cm}<{\centering} m{0.6cm}<{\centering} m{6.5cm}<{\centering}} \cline{1-7}
		$q$ &$q_{sub1}$ &$q_{sub2}$ &DFGN &DecompRC &HGN &PCL &\multirow{9}{*}{\begin{minipage}[l]{5\columnwidth}
		\raisebox{-.5\height}{\includegraphics[scale=0.32]{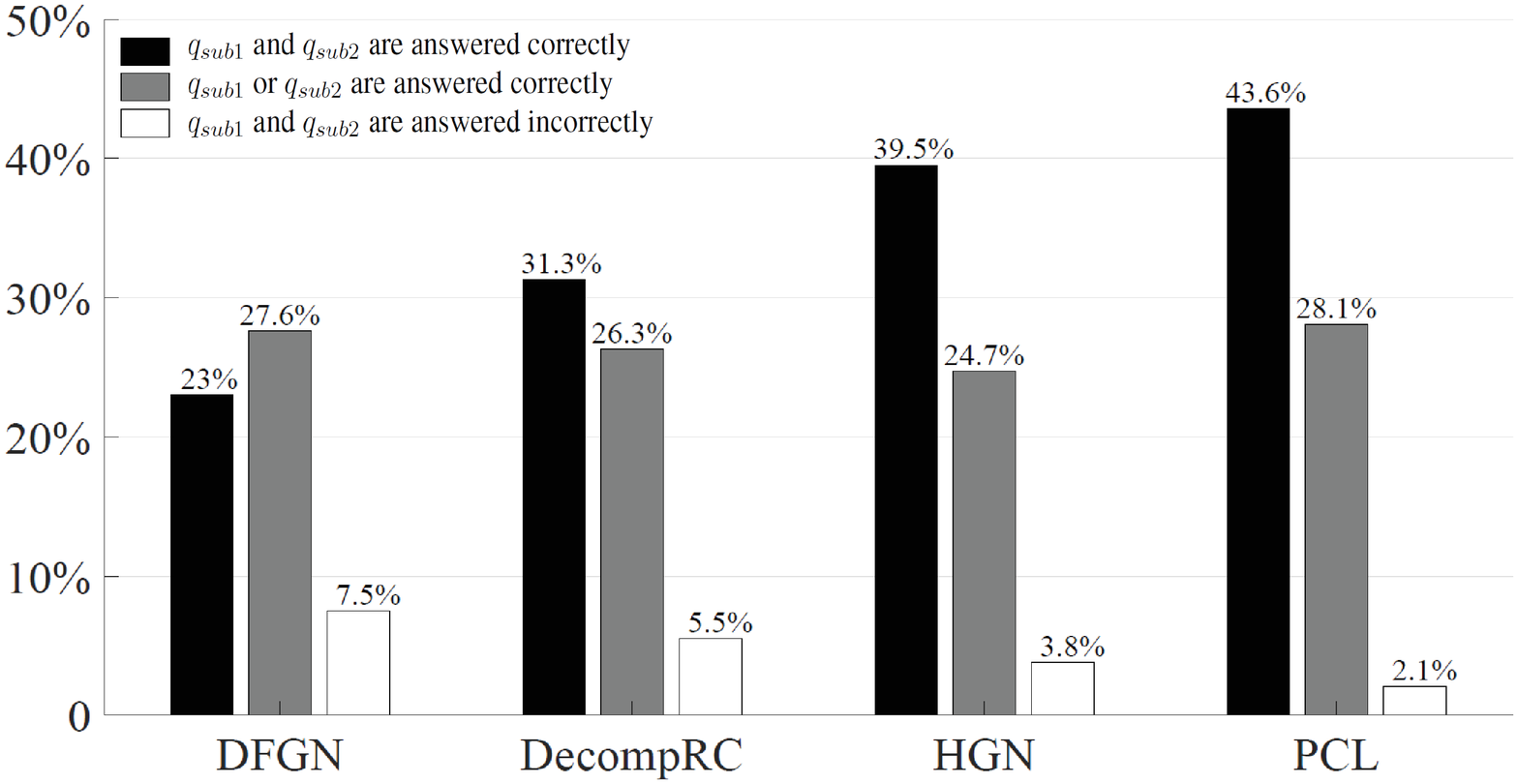}}
	\end{minipage}} \\ \cline{1-7}	
		c 	&c	&c  &23    &31.3	&39.5  &43.6\\ \cline{1-7}			
        c 	&c	&w	&9.7   &7.2  	&5.1   &6.8 \\ \cline{1-7}		
        c 	&w	&c  &17.9  &19.1	&19.6  &21.3 \\	\cline{1-7}		
        c 	&w	&w  &7.5   &5.5 	&3.8   &2.1 \\	\cline{1-7}			
        w	&c	&c  &4.9   &3	    &2.8   &1.7 \\	\cline{1-7}			
        w	&c	&w  &17	   &18.6	&16.7  &16.3 \\	\cline{1-7}		
        w	&w	&c  &3.5   &3.4     &2.6   &1.1 \\	\cline{1-7}		
        w	&w	&w  &16.5  &11.9	&9.9   &7.1 \\  \cline{1-7}
	\end{tabular}
	\vspace{-2mm}
	\caption{\footnotesize (Left) Categorical EM statistics (\%) of sub-question evaluation for four multi-hop QA models. $c/w$ denotes that the question is answered correctly/wrongly. For example, the first four rows show the percentage of multi-hop questions that can be correctly answered. (Right) The success rate of four multi-hop QA models.}
	\label{fig3}
\end{table*}

\subsection{Ablation Studies}
To verify the effect of the components in our PCL model, we perform the following ablation studies on the dev set of HotpotQA.

\paragraph{Effect of Conservation Learning (CL).} To verify the effect of conservation learning on multi-hop QA, we compare performance with the PLM ELECTRA with and without conservation learning. For conservation learning, we first trained an ELECTRA-based QA model on the single-hop QA dataset SQuAD \cite{rajpurkar2016squad}, and then retrained it on the HotpotQA dataset with conservation learning. As shown in Table \ref{tab2}, we observe that the overall performance (F1 score) increased from 73.89 to 76.39 after using conservation learning, which shows that our model performs well on multi-hop reasoning when the previously learned knowledge is retained. In the following Section 4.6, we provide an in-depth analysis on the performance of our model on the sub-questions, to compare the ability of models to mitigate forgetting.

\begin{table}
	\small
	\centering
	\begin{tabular} {l c c c} \hline
        Model	        &Ans F1	&Sup F1	&Joint F1 \\ \hline
		ELECTRA       &81.05  &89.97  &73.89  \\
        - Prompt	  &82.06  &90.36  &75.02 \\ 
        - CL	      &82.99  &90.97  &76.39 \\ \hline
        PCL           &\textbf{84.42}  &\textbf{91.15}  &\textbf{77.76} \\ \hline
	\end{tabular}
	\caption{\footnotesize Ablation Study of PCL on the dev set of HotpotQA. Prompt denotes that a soft prompt is used to condition PLM ELECTRA to stimulate the reasoning required for the multi-hop question. CL denotes that conservation learning is used to perform multi-hop reasoning. PCL used both soft prompts and conservation learning.}
 	\vspace{-3mm}
	\label{tab2}
\end{table}

\paragraph{Effect of Soft Prompts.} To verify the effect of the soft prompt and perform type-specific reasoning, we first identified the reasoning type of the multi-hop question using a classifier based on ELECTRA. In Table \ref{tab3}, our classifier QC$_{\rm ELECTRA}$ achieves good accuracy compared to DecompRC, providing a solid basis for type-specific multi-hop reasoning. Then, we transform the identified reasoning type into a soft prompt to stimulate the PLM to perform the corresponding type of multi-hop reasoning. In Table \ref{tab2}, we implant the soft prompt both into the baseline ELECTRA and the ELECTRA based on conservation learning (PCL), the Joint F1 score improved by 1.23 and 1.37, respectively. This suggests that the soft prompt based on the reasoning type can stimulate the question-type-specific reasoning knowledge required for multi-hop QA.

\begin{table}
	\small
	\centering
	\begin{tabular} {l c} \hline
    Model	&Accuracy\\ \hline
    DecompRC           &70.40	\\	
    QC(ELECTRA-large) 		    &\textbf{98.97} \\ \hline
	\end{tabular}
	\caption{\footnotesize The performance of question classification by different models. QC(ELECTRA-large) is a question classifier based on ELECTRA-large.}
	\label{tab3}
	\vspace{-3mm}
\end{table}

\paragraph{Effect of Pre-trained Language Model.} To verify the effects of PLMs, we compare PCL with HGN based on the same data and backbone. As shown in Table \ref{tab5}, PCL outperforms HGN on all metrics. This indicates the effectiveness and robustness of PCL across PLMs. 

\begin{table}
	\small
	\centering
	\begin{tabular} {l c c c} \hline
        Model	        &Ans F1	&Sup F1	&Joint F1 \\ \hline
        HGN(RoBERTa)	&82.22	&88.58	&74.37 \\
        HGN(ELECTRA)	&82.24	&88.63	&74.51 \\
        HGN(ALBERT)	    &83.46	&89.2	&75.79 \\ \hline
        PCL(RoBERTa)	&84.33	&90.75	&77.12 \\
        PCL(ELECTRA)	&84.42	&91.15	&77.76 \\  
        PCL(ALBERT)	    &85.47	&91.28	&78.76 \\ \hline
	\end{tabular}
	\caption{\footnotesize Results with different PLMs on the dev set of HotpotQA. RoBERTa, ELECTRA and ALBERT denote that we use RoBERTa-large, ELECTRA-large and ALBERT-xxlarge-v2 as the PLM respectively.}
	\label{tab5}
\end{table}

\subsection{Evaluation across Reasoning Types}
We evaluate the performance of PCL for multi-hop questions with multiple reasoning types. Specifically, we follow HGN in splitting the multi-hop questions into three categories: bridge, comparison-yes/no and comparison-span. ``Bridge'' questions require identifying a bridge entity to infer the answer, ``comparison-yes/no'' and ``comparison-span'' require comparing two entities to infer the answer that could be yes/no or a span of text. As shown in Table \ref{tab4}, our PCL performs better than HGN for all reasoning types, indicating that the performance of the model can be effectively improved by using soft prompts for type-specific reasoning. 

\begin{table}
	\small
	\centering
	\begin{tabular} {l c c c c} \hline
        Model	&Question &Ans F1	&Sup F1	&Joint F1 \\ \hline
                          & bridge	  &81.90	    &87.60	&73.31 \\ 
        HGN                  & comp-yn    &93.45	&94.22	&88.5  \\
                          & comp-span  &79.06	&91.72	&74.17 \\ \hline
                          & bridge	  &85.36	&90.77	&78.17 \\ 
        PCL                  & comp-yn    &93.67	&94.73	&88.93  \\
                          & comp-span  &82.42	&92.65	&77.57 \\ \hline
	\end{tabular}
	\caption{\footnotesize Results with different reasoning types on the dev set of HotpotQA. PCL outperforms HGN in all reasoning types.}
	\label{tab4}
	\vspace{-3mm}
\end{table}


\subsection{Evaluation of Robustness}
In this section, we evaluate the robustness and generalization of PCL on three different datasets. 

\paragraph{Evaluation on Sub-question Dataset.} To analyze whether existing multi-hop QA models could at least in principle perform the multi-hop reasoning process by composing an answer out of solved sub-questions, we perform an evaluation on 1000 human-verified examples \cite{tang-etal-2021-multi}. These data  consist of 1000 multi-hop questions $q$, and the corresponding 1000 sub-questions $q_{sub1}$, $q_{sub2}$. EM and F1 are used in each case to evaluate performance on answer prediction. As shown in Table \ref{tab6}, PCL achieves the best performance on the 1000 human-verified examples. Compared to DFGN and DecompRC, whose performance significantly drops on sub-questions, especially on the second sub-questions. PCL dropped by only 2.4 on average, which demonstrates that PCL can in principle support the expected behaviour on each hop of the reasoning process better than other multi-hop QA models by mitigating knowledge forgetting. 

\begin{table}
	\small
	\centering
	\begin{tabular} {l c c c c c c} \hline
		\multirow{2}{*}{Model}  &\multicolumn{2}{c}{$q$} &\multicolumn{2}{c}{$q_{sub1}$} &\multicolumn{2}{c}{$q_{sub2}$} \\ \cmidrule(lr){2-3} \cmidrule(lr){4-5} \cmidrule(lr){6-7} 
		&EM &F1 &EM &F1 &EM &F1  \\\hline
		DFGN     &58.1	&71.96	&54.6	&68.54	&49.3	&60.83  \\
		DecRC    &63.1	&77.61	&61.0	    &75.21	&56.8	&70.77 \\
        HGN      &71.0	&84.25	&66.1	&81.72	&66.7	&78.24 \\ \hline
        PCL     &\textbf{73.8}  &\textbf{87.15}  &\textbf{68.4}   &\textbf{83.62}  &\textbf{68.5}  &\textbf{81.07} \\ \hline
	\end{tabular}
	\caption{\footnotesize Results on the sub-question dataset with different multi-hop QA models. $q$ denotes the multi-hop question, $q_{sub1}$ and $q_{sub2}$ denote the corresponding sub-questions of $q$.}
	\vspace{-2mm}
	\label{tab6}
\end{table}

\begin{table}
	\small
	\centering
	\begin{tabular} {l c c c c } \hline
		Train & \multicolumn{2}{c}{Reg} & \multicolumn{2}{c}{Reg} \\ 
		Eval  & \multicolumn{2}{c}{Reg} & \multicolumn{2}{c}{Adv} \\  \hline
		Model &EM &F1 &EM &F1 \\\hline
		HGN         &47.31 &74.37 &41.56 &69.81 \\
		PCL         &49.59 &77.76 &47.87 &74.24	 \\ \hline
	\end{tabular}
	\caption{\footnotesize EM and F1 scores after evaluating on the adversarial dataset designed to probe for the use of unsound reasoning shortcuts. Reg or Adv denotes training or evaluating the model on the standard or adversarial HotpotQA dataset.}
	\label{tab7}
	\vspace{-3mm}
\end{table}

To further analyze whether models effectively mitigate knowledge forgetting, we collect the correctness statistics on each example in the sub-question dataset. As shown in Table \ref{fig3} (Left), PCL has a 96.25\% chance of getting the parent multi-hop question $q$ right when both sub-questions $q_{sub1}$ and $q_{sub2}$ are answered correctly, which indicates that PCL can better retain the learned knowledge, through its use of conservation learning, compared with other multi-hop QA models. However, we observe that PCL still has a high probability of answering the parent multi-hop question correctly when only one of the sub-question is answered correctly. We summarize the sub-question dependent success rate of multi-hop QA models in Table \ref{fig3} (Right). We observe that these models can answer parent multi-hop questions with a high probability (exceeding 20\%) when only one sub-question is answered correctly, which indicates that using potentially unsound reasoning shortcuts to predict answers is a common and difficult to avoid phenomenon in multi-hop QA. 

\paragraph{Evaluation on Adversarial Dataset.} To compare the extent to which models are currently able to avoid the unsound-reasoning-shortcut problem, we conducted an adversarial evaluation on the dev set of HotpotQA, reported in  Table \ref{tab7}. In the adversarial examples, the fake answers are sampled from the original HotpotQA dataset, but do not affect the validity of the original answers. As shown in Table \ref{tab7}, we trained PCL and HGN on the standard training data and evaluated them on both the standard and adversarial dev data. The result shows that PCL achieves better performance than HGN, indicating that PCL is more robust than HGN against the use of shortcuts probed by the adversarial dataset.

\begin{table}
	\small
	\centering
	\begin{tabular} {l c c c c} \hline
		\multirow{2}{*}{}  &\multicolumn{2}{c}{2WikiMultihopQA} &\multicolumn{2}{c}{MusiQue} \\ \cmidrule(lr){2-3} \cmidrule(lr){4-5}
		&EM &F1 &EM &F1 \\\hline
		HGN    &38.74 &68.69 &39.42 &65.12 \\
		PCL    &46.03 &73.42 &41.28 &67.34	 \\ \hline
	\end{tabular}
	\caption{\footnotesize Results of PCL and HGN on 2WikiMultihopQA and MusiQue multi-hop QA dataset.}
	\vspace{-1mm}
	\label{tab8}
\end{table}

\begin{figure*}[!ht]
	\begin{center}
		\vspace{-3mm}
		\subfigure{\scalebox{0.5} {\includegraphics{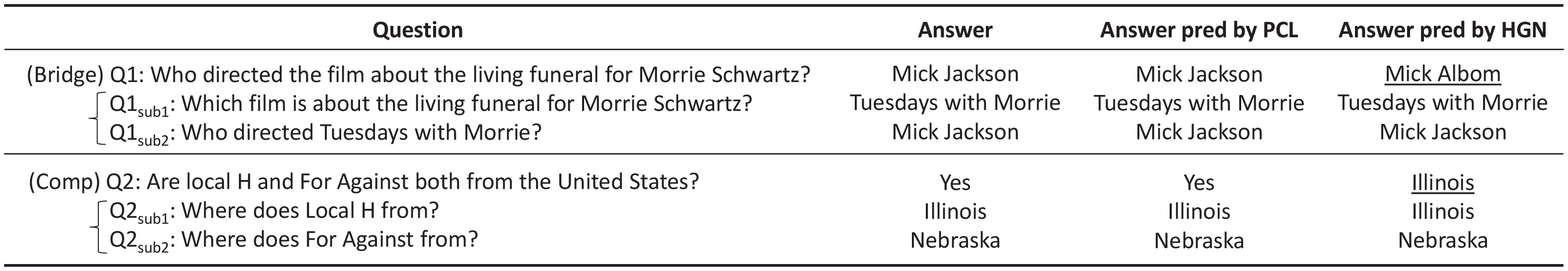}}}
		\subfigure{\scalebox{0.5} {\includegraphics{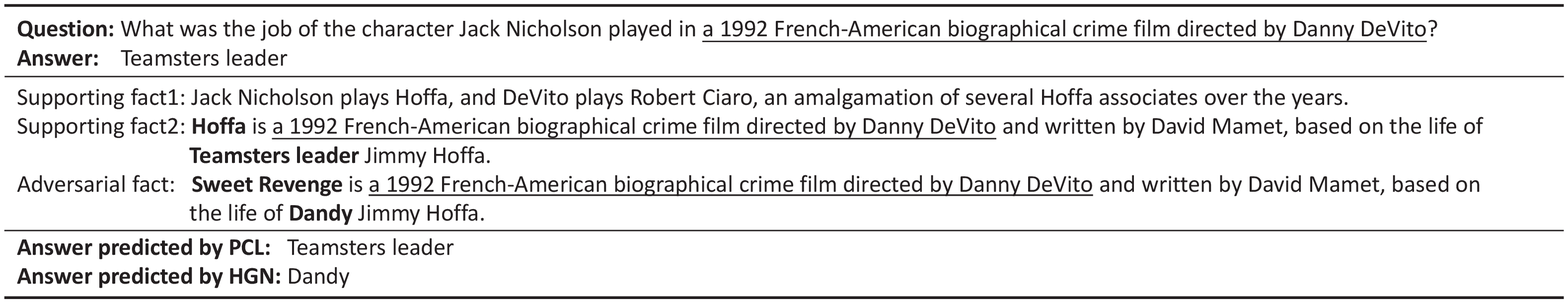}}}
		\vspace{-4mm}
		\caption{\footnotesize Case studies of the sub-question evaluation and adversarial multi-hop question evaluation. The upper case study indicates that our PCL has stronger composite reasoning ability compared to HGN. The lower case study indicates that the iterative paragraph selection is help to avoid predict the answer by using reasoning shortcuts.}
		\label{fig4}
		\vspace{-1mm}
	\end{center}
\end{figure*}

\paragraph{Evaluation on Other Multi-hop Datasets.} To verify whether PCL can generalize to other multi-hop QA datasets, we compared PCL against HGN on the 2WikiMultihopQA and MuSiQue dataset. In Table \ref{tab8} we observe that PCL outperforms HGN on these two datasets, which demonstrates PCL's good potential on generalisation to QA problems with more than 2 hops. 


\subsection{Case Study}
We present two case studies in Figure \ref{fig4}. The upper case illustrates the results of PCL and HGN at each hop of the reasoning process. We observe that PCL correctly answered the bridge question Q1, while HGN did not, when all sub-questions were answered correctly,  supporting the claim that PCL learns new QA knowledge while retaining knowledge learned for sub-questions. Similarly, for comparison question Q2, PCL learned the specific reasoning ability based on the reasoning type to which Q2 belongs, indicating soft prompts based on reasoning types can elicit the reasoning knowledge required for multi-hop questions. 

The lower case illustrates the results of PCL and HGN on an adversarial multi-hop question. In the example, the question can be directly answered by matching a reasoning shortcut in supporting facts2 \textit{``a 1992 French-American biographical crime film directed by Danny DeVito''}. To avoid it, we follow \cite{jiang2019avoiding} to construct an adversarial fact from the candidate paragraphs by replacing the subject and the answer, \eg ``\textit{Sweet Revenge}'' for ``\textit{Hoffa}''  and ``\textit{Dandy}'' for ``\textit{Teamsters leader}''. We observed that PCL correctly answered the question despite the interference from the adversarial fact, while HGN did not. This supports the claim that the iterative paragraph selection helps establish connections between supporting facts, because PCL selects the next supporting fact2 based on the previous supporting fact1. In this example, the adversarial fact is irrelevant to supporting fact1, so PCL excludes it during paragraph selection.

\section{Conclusions and Future Work}
In this paper, we introduce a novel prompt-based conservation learning framework for multi-hop QA -- a framework  that retains knowledge from previous component tasks -- able to answer questions in a principled way that matches human expectations by answering sub-questions and integrating the answers.
By developing soft prompts related to reasoning types during training, we also show that we can condition PLMs to stimulate and apply the reasoning knowledge required for specific multi-hop questions. Experimental results on multiple multi-hop QA datasets demonstrate the improved performance of PCL over previous multi-hop QA models in multi-hop QA.

Next, we plan to extend PCL on QA problems with arbitrary hop-counts, and
 to increase generality by extending soft prompts to handle QA with unrestricted numbers of, and implicit, reasoning types, and non-linear reasoning structures.



\bibliography{custom}

\begin{thebibliography}{35}
\expandafter\ifx\csname natexlab\endcsname\relax\def\natexlab#1{#1}\fi

\bibitem[{Beltagy et~al.(2020)Beltagy, Peters, and
  Cohan}]{beltagy2020longformer}
Iz~Beltagy, Matthew~E Peters, and Arman Cohan. 2020.
\newblock Longformer: The long-document transformer.
\newblock \emph{arXiv preprint arXiv:2004.05150}.

\bibitem[{Clark et~al.(2020)Clark, Luong, Le, and Manning}]{clark2020electra}
Kevin Clark, Minh-Thang Luong, Quoc~V Le, and Christopher~D Manning. 2020.
\newblock Electra: Pre-training text encoders as discriminators rather than
  generators.
\newblock \emph{arXiv preprint arXiv:2003.10555}.

\bibitem[{Fang et~al.(2020)Fang, Sun, Gan, Pillai, Wang, and
  Liu}]{fang2019hierarchical}
Yuwei Fang, Siqi Sun, Zhe Gan, Rohit Pillai, Shuohang Wang, and Jingjing Liu.
  2020.
\newblock Hierarchical graph network for multi-hop question answering.
\newblock In \emph{EMNLP}, pages 8823--8838.

\bibitem[{Gu et~al.(2021)Gu, Han, Liu, and Huang}]{gu2021ppt}
Yuxian Gu, Xu~Han, Zhiyuan Liu, and Minlie Huang. 2021.
\newblock Ppt: Pre-trained prompt tuning for few-shot learning.
\newblock \emph{arXiv preprint arXiv:2109.04332}.

\bibitem[{Ho et~al.(2020)Ho, Nguyen, Sugawara, and Aizawa}]{ho2020constructing}
Xanh Ho, Anh-Khoa~Duong Nguyen, Saku Sugawara, and Akiko Aizawa. 2020.
\newblock Constructing a multi-hop qa dataset for comprehensive evaluation of
  reasoning steps.
\newblock \emph{arXiv preprint arXiv:2011.01060}.

\bibitem[{Jiang and Bansal(2019)}]{jiang2019avoiding}
Yichen Jiang and Mohit Bansal. 2019.
\newblock Avoiding reasoning shortcuts: Adversarial evaluation, training, and
  model development for multi-hop qa.
\newblock \emph{arXiv preprint arXiv:1906.07132}.

\bibitem[{Li et~al.(2021{\natexlab{a}})Li, Krishnan, Wu, Kolouri, Pilly, and
  Braverman}]{li2021lifelong}
Haoran Li, Aditya Krishnan, Jingfeng Wu, Soheil Kolouri, Praveen~K Pilly, and
  Vladimir Braverman. 2021{\natexlab{a}}.
\newblock Lifelong learning with sketched structural regularization.
\newblock In \emph{ACML}, pages 985--1000. PMLR.

\bibitem[{Li et~al.(2021{\natexlab{b}})Li, Wang, Wang, and
  Jiang}]{liasynchronous}
Ronghan Li, Lifang Wang, Shengli Wang, and Zejun Jiang. 2021{\natexlab{b}}.
\newblock Asynchronous multi-grained graph network for interpretable multi-hop
  reading comprehension.
\newblock In \emph{IJCAI}, pages 3857--3863.

\bibitem[{Li and Liang(2021)}]{li2021prefix}
Xiang~Lisa Li and Percy Liang. 2021.
\newblock Prefix-tuning: Optimizing continuous prompts for generation.
\newblock \emph{arXiv preprint arXiv:2101.00190}.

\bibitem[{Liu et~al.(2021)Liu, Yuan, Fu, Jiang, Hayashi, and
  Neubig}]{liu2021pre}
Pengfei Liu, Weizhe Yuan, Jinlan Fu, Zhengbao Jiang, Hiroaki Hayashi, and
  Graham Neubig. 2021.
\newblock Pre-train, prompt, and predict: A systematic survey of prompting
  methods in natural language processing.
\newblock \emph{arXiv preprint arXiv:2107.13586}.

\bibitem[{Min et~al.(2019)Min, Zhong, Zettlemoyer, and
  Hajishirzi}]{min2019multi}
Sewon Min, Victor Zhong, Luke Zettlemoyer, and Hannaneh Hajishirzi. 2019.
\newblock Multi-hop reading comprehension through question decomposition and
  rescoring.
\newblock \emph{arXiv preprint arXiv:1906.02916}.

\bibitem[{Nishida et~al.(2019)Nishida, Nishida, Nagata, Otsuka, Saito, Asano,
  and Tomita}]{nishida2019answering}
Kosuke Nishida, Kyosuke Nishida, Masaaki Nagata, Atsushi Otsuka, Itsumi Saito,
  Hisako Asano, and Junji Tomita. 2019.
\newblock Answering while summarizing: Multi-task learning for multi-hop qa
  with evidence extraction.
\newblock \emph{arXiv preprint arXiv:1905.08511}.

\bibitem[{Parisi et~al.(2019)Parisi, Kemker, Part, Kanan, and
  Wermter}]{parisi2019continual}
German~I Parisi, Ronald Kemker, Jose~L Part, Christopher Kanan, and Stefan
  Wermter. 2019.
\newblock Continual lifelong learning with neural networks: A review.
\newblock \emph{Neural Networks}, 113:54--71.

\bibitem[{Perez et~al.(2020)Perez, Lewis, Yih, Cho, and
  Kiela}]{perez2020unsupervised}
Ethan Perez, Patrick S.~H. Lewis, Wen{-}tau Yih, Kyunghyun Cho, and Douwe
  Kiela. 2020.
\newblock Unsupervised question decomposition for question answering.
\newblock In \emph{EMNLP}, pages 8864--8880.

\bibitem[{Qi et~al.(2020)Qi, Lee, Sido, Manning et~al.}]{qi2020answering}
Peng Qi, Haejun Lee, Oghenetegiri Sido, Christopher~D Manning, et~al. 2020.
\newblock Answering open-domain questions of varying reasoning steps from text.
\newblock \emph{arXiv preprint arXiv:2010.12527}.

\bibitem[{Qi et~al.(2019)Qi, Lin, Mehr, Wang, and Manning}]{qi2019answering}
Peng Qi, Xiaowen Lin, Leo Mehr, Zijian Wang, and Christopher~D Manning. 2019.
\newblock Answering complex open-domain questions through iterative query
  generation.
\newblock \emph{arXiv preprint arXiv:1910.07000}.

\bibitem[{Qin et~al.(2022)Qin, Zhang, Lin, Liu, Li, Sun, and
  Zhou}]{qin2022elle}
Yujia Qin, Jiajie Zhang, Yankai Lin, Zhiyuan Liu, Peng Li, Maosong Sun, and Jie
  Zhou. 2022.
\newblock Elle: Efficient lifelong pre-training for emerging data.
\newblock \emph{arXiv preprint arXiv:2203.06311}.

\bibitem[{Qiu et~al.(2019)Qiu, Xiao, Qu, Zhou, Li, Zhang, and
  Yu}]{qiu2019dynamically}
Lin Qiu, Yunxuan Xiao, Yanru Qu, Hao Zhou, Lei Li, Weinan Zhang, and Yong Yu.
  2019.
\newblock Dynamically fused graph network for multi-hop reasoning.
\newblock In \emph{ACL}, pages 6140--6150.

\bibitem[{Rajpurkar et~al.(2016)Rajpurkar, Zhang, Lopyrev, and
  Liang}]{rajpurkar2016squad}
Pranav Rajpurkar, Jian Zhang, Konstantin Lopyrev, and Percy Liang. 2016.
\newblock Squad: 100,000+ questions for machine comprehension of text.
\newblock \emph{arXiv preprint arXiv:1606.05250}.

\bibitem[{Rolnick et~al.(2019)Rolnick, Ahuja, Schwarz, Lillicrap, and
  Wayne}]{rolnick2019experience}
David Rolnick, Arun Ahuja, Jonathan Schwarz, Timothy Lillicrap, and Gregory
  Wayne. 2019.
\newblock Experience replay for continual learning.
\newblock \emph{Advances in Neural Information Processing Systems}, 32.

\bibitem[{Saxena et~al.(2020)Saxena, Tripathi, and
  Talukdar}]{saxena2020improving}
Apoorv Saxena, Aditay Tripathi, and Partha Talukdar. 2020.
\newblock Improving multi-hop question answering over knowledge graphs using
  knowledge base embeddings.
\newblock In \emph{ACL}, pages 4498--4507.

\bibitem[{Schwartz et~al.(2020)Schwartz, Dodge, Smith, and
  Etzioni}]{schwartz2020green}
Roy Schwartz, Jesse Dodge, Noah~A Smith, and Oren Etzioni. 2020.
\newblock Green ai.
\newblock \emph{Communications of the ACM}, 63(12):54--63.

\bibitem[{Shao et~al.(2020{\natexlab{a}})Shao, Cui, Liu, Wang, and
  Hu}]{shao-etal-2020-graph}
Nan Shao, Yiming Cui, Ting Liu, Shijin Wang, and Guoping Hu.
  2020{\natexlab{a}}.
\newblock \href {https://doi.org/10.18653/v1/2020.emnlp-main.583} {Is {G}raph
  {S}tructure {N}ecessary for {M}ulti-hop {Q}uestion {A}nswering?}
\newblock In \emph{EMNLP}, pages 7187--7192, Online. Association for
  Computational Linguistics.

\bibitem[{Shao et~al.(2020{\natexlab{b}})Shao, Cui, Liu, Wang, and
  Hu}]{shao2020graph}
Nan Shao, Yiming Cui, Ting Liu, Shijin Wang, and Guoping Hu.
  2020{\natexlab{b}}.
\newblock Is graph structure necessary for multi-hop question answering?
\newblock In \emph{EMNLP}, pages 7187--7192.

\bibitem[{Sun et~al.(2019)Sun, Ho, and Lee}]{sun2019lamol}
Fan-Keng Sun, Cheng-Hao Ho, and Hung-Yi Lee. 2019.
\newblock Lamol: Language modeling for lifelong language learning.
\newblock \emph{arXiv preprint arXiv:1909.03329}.

\bibitem[{Sun et~al.(2020)Sun, Wang, Li, Feng, Tian, Wu, and
  Wang}]{sun2020ernie}
Yu~Sun, Shuohuan Wang, Yukun Li, Shikun Feng, Hao Tian, Hua Wu, and Haifeng
  Wang. 2020.
\newblock Ernie 2.0: A continual pre-training framework for language
  understanding.
\newblock In \emph{AAAI}.

\bibitem[{Talmor and Berant(2018)}]{talmor2018web}
Alon Talmor and Jonathan Berant. 2018.
\newblock The web as a knowledge-base for answering complex questions.
\newblock \emph{arXiv preprint arXiv:1803.06643}.

\bibitem[{Tang et~al.(2021)Tang, Ng, and Tung}]{tang-etal-2021-multi}
Yixuan Tang, Hwee~Tou Ng, and Anthony Tung. 2021.
\newblock \href {https://doi.org/10.18653/v1/2021.eacl-main.283} {Do multi-hop
  question answering systems know how to answer the single-hop sub-questions?}
\newblock In \emph{ACL}, pages 3244--3249, Online. Association for
  Computational Linguistics.

\bibitem[{Trivedi et~al.(2021)Trivedi, Balasubramanian, Khot, and
  Sabharwal}]{trivedi2021musique}
Harsh Trivedi, Niranjan Balasubramanian, Tushar Khot, and Ashish Sabharwal.
  2021.
\newblock Musique: Multi-hop questions via single-hop question composition.
\newblock \emph{arXiv preprint arXiv:2108.00573}.

\bibitem[{Tu et~al.(2020)Tu, Huang, Wang, Huang, He, and Zhou}]{tu2020select}
Ming Tu, Kevin Huang, Guangtao Wang, Jing Huang, Xiaodong He, and Bowen Zhou.
  2020.
\newblock Select, answer and explain: Interpretable multi-hop reading
  comprehension over multiple documents.
\newblock In \emph{AAAI}, pages 9073--9080.

\bibitem[{Welbl et~al.(2018)Welbl, Stenetorp, and
  Riedel}]{welbl2018constructing}
Johannes Welbl, Pontus Stenetorp, and Sebastian Riedel. 2018.
\newblock Constructing datasets for multi-hop reading comprehension across
  documents.
\newblock \emph{Transactions of the Association for Computational Linguistics},
  6:287--302.

\bibitem[{Wolf et~al.(2020)Wolf, Debut, Sanh, Chaumond, Delangue, Moi, Cistac,
  Rault, Louf, Funtowicz et~al.}]{wolf2020transformers}
Thomas Wolf, Lysandre Debut, Victor Sanh, Julien Chaumond, Clement Delangue,
  Anthony Moi, Pierric Cistac, Tim Rault, R{\'e}mi Louf, Morgan Funtowicz,
  et~al. 2020.
\newblock Transformers: State-of-the-art natural language processing.
\newblock In \emph{EMNLP}, pages 38--45.

\bibitem[{Wu et~al.(2021)Wu, Zhang, and Zhao}]{wu2021graph}
Bohong Wu, Zhuosheng Zhang, and Hai Zhao. 2021.
\newblock Graph-free multi-hop reading comprehension: A select-to-guide
  strategy.
\newblock \emph{arXiv preprint arXiv:2107.11823}.

\bibitem[{Yang et~al.(2018)Yang, Qi, Zhang, Bengio, Cohen, Salakhutdinov, and
  Manning}]{yang-etal-2018-hotpotqa}
Zhilin Yang, Peng Qi, Saizheng Zhang, Yoshua Bengio, William Cohen, Ruslan
  Salakhutdinov, and Christopher~D. Manning. 2018.
\newblock \href {https://doi.org/10.18653/v1/D18-1259} {{H}otpot{QA}: A dataset
  for diverse, explainable multi-hop question answering}.
\newblock In \emph{Proceedings of the 2018 Conference on Empirical Methods in
  Natural Language Processing}, pages 2369--2380, Brussels, Belgium.
  Association for Computational Linguistics.

\bibitem[{Zhu et~al.(2021)Zhu, Pang, Lan, Shen, and Cheng}]{zhu2021adaptive}
Yunchang Zhu, Liang Pang, Yanyan Lan, Huawei Shen, and Xueqi Cheng. 2021.
\newblock Adaptive information seeking for open-domain question answering.
\newblock \emph{arXiv preprint arXiv:2109.06747}.

\end{thebibliography}

\clearpage
\end{document}